\documentclass{article}[10pt]

\title{\bf Transforming Business Rules into Natural Language Text}
\author
{Manuela Kunze and Dietmar R{\"o}sner
\\ Otto-von-Guericke-Universit{\"a}t Magdeburg
\\ Institut f{\"u}r Wissens- und Sprachverarbeitung
\\ P.O. Box 4120
\\ D--39016 Magdeburg, Germany,
\\ makunze$|$roesner@iws.cs.uni-magdeburg.de}

\begin{document}
\maketitle

\begin{abstract}
The aim of the project presented in this paper is to design a
system for an NLG\footnote{Natural Language Generation}
architecture, which supports the documentation process of
eBusiness models. A major task is to enrich the formal description
of an eBusiness model with additional information needed in an NLG
task.
\end{abstract}

\subsection*{Introduction} Communication and explanation of
business rules defined in an eBusiness model is a crucial point
during the design process of new electronic market places. The
documentation of these complex models is an expensive process. The
conceptualization of eBusiness models will be realized in a formal
way. The transformation of the conceptualization language into an
agent language for software agents is easier than to communicate
the rules of the market place to a human user.

The manual documentation process of an eBusiness model is time and
cost consuming. Furthermore, it can not be guaranteed that the
documents created manually are complete and understandable for
every reader.

The main task of the project `\emph{Formalization and
Documentation of Electronic Market Place Rules of Real Internet
Platforms}' (FormDoc) is to investigate how a formal description
of an eBusiness platform must and can be enriched for
the transformation into natural language text. 

\subsection*{The FormDoc Project} The tasks of the project
can be divided into three subtasks: i. to analyse the language
used in real eBusiness documents ii. to investigate current
representations formats for business rules (e.g. CLP
\cite{GroLaCha:2002}), and iii. to specify the requirements for an
NLG architecture for the documentation task.

Crucial questions of the first two subtasks are:
\begin{itemize}
    \item In which ways can formal concepts be expressed by natural language
    terms?
    \item Which rules and concepts can be summarized? Which linguistic
    means can be used for aggregation of rules?
    \item Which concepts and rules must be verbalized in the
    documents?
\end{itemize}
\normalsize

%

It is necessary to carry out experiments with different languages
(in our case German and English), because eBusiness market places
are available in WWW. One initial point for the last subtask is to
test the KPML environment \cite{Bateman95-enlgw}. KPML contains
different language resources, which can be easily extended and
adapted for our requirements. But the crucial question is: Is the
necessary linguistic competence available in the front end
generator KPML?


In the project outline above, a main task is the design of an
appropriate semantic representation for business rules. The
semantic representation should be useable for
\begin{itemize}
    \item implementing the market place,
    \item simulating the market place, and
    \item generating natural language documentation about the
    rules of the market place.
\end{itemize}

For the generation process, it is necessary to augment the rules
within an eBusiness model with linguistic information (like canned
phrases, linguistic terms, synonyms, etc.). Another main focus for
the generation process is to group or to summarize the rules,
because in order to avoid repetition and verbosity, the rules
within a model should be grouped into clusters of rules that
describe a similar content. For example, following rules are
given:

\begin{verbatim}
<steadySpender>
  if shopper(?Cust) and  spendingHistory(?Cust, loyal)
  then giveDiscount(percent5, ?Cust);

<platinumClub>
   if shopper(?Cust) and  memberPlatinumClub(?Cust)
   then giveDiscount(percent10, ?Cust);
\end{verbatim}

These two rules can be expressed with the following single
sentence: \emph{Customers with a loyal spending history obtain a
discount of 5 \% but as member of a platinum card you obtain a
discount of 10
  \%.}

Another kind of summarizing can be realized for the next example:
\begin{verbatim}
<steadySpender>
  if shopper(?Cust) and  spendingHistory(?Cust, loyal)
  then giveDiscount(percent5, ?Cust);

<storeCard>
   if shopper(?Cust) and hasChargeCard(?Cust, store)
   then giveDiscount(percent5, ?Cust);
\end{verbatim}

In this case, the same implication is given but different premises
are described. These rules can be summarized through the following
sentence: \emph{Customers with a loyal spending history or members
of our charge card obtain a discount of 5 \%.}

The work of the project started with a feasibility study using
KPML as NLG environment and first analyses of the characteristics
of the sublanguage used in real eBusiness documents. Currently,
the project work is concentrated on the description of a (meta
data) model for the interpretation and selection of business
rules. This model should be used to bridge the gap between a
formal semantic representation and natural language text. The
model describes which additional information or knowledge about
rules in an eBusiness model is necessary for the NLG process.

\bibliographystyle{alpha}
\bibliography{iwcs-project}

\begin{thebibliography}{GLC99}

\bibitem[Bat95]{Bateman95-enlgw}
J.~A. Bateman.
\newblock {KPML: The KOMET-Penman Multilingual Resource Development
  Environment}.
\newblock In {\em Proceedings of the Fifth European Workshop on Natural
  Language Generation}, pages 219--222, Leiden, The Netherlands, May 1995.
  Faculty of Social and Behavioural Sciences, University of Leiden.

\bibitem[GLC99]{GroLaCha:2002}
Benjamin~N. Grosof, Yannis Labrou, and Hoi~Y. Chan.
\newblock {A Declarative Approach to Business Rules in Contracts: Courteous
  Logic Programs in XML}.
\newblock In {\em {Proceedings of the First ACM Conference on Electronic
  Commerce (EC-99)}}. {Denver, CO}, 1999.

\end{thebibliography}

\end{document}